\documentclass{article}

\usepackage{arxiv}

\usepackage{newtxtext}
\usepackage{newtxmath}

\usepackage[utf8]{inputenc}
\usepackage[T2A,T1]{fontenc}
\usepackage[english]{babel}

\usepackage{hyperref}
\usepackage{url}
\usepackage{booktabs}
\usepackage{amsfonts}
\usepackage{nicefrac}
\usepackage{microtype}
\usepackage{cleveref}
\usepackage{graphicx}
\usepackage{natbib}
\usepackage{doi}
\usepackage{authblk}

\usepackage{orcidlink}

\title{Benchmarking Arabic--Russian Machine Translation: A Comparison of Fine-tuned NMT and Few-shot LLMs under Rich Morphology and Low Lexical Overlap}

\author{Mullosharaf~K.~Arabov\thanks{Email: \texttt{MKArabov@kpfu.ru}}\orcidlink{0000-0003-2525-1183}}
\affil{Kazan Federal University, Institute of Computational Mathematics and Information Technologies, Kazan, Russia}

\renewcommand{\shorttitle}{Arabic--Russian MT: NMT vs Few-shot LLMs}

\hypersetup{
    pdftitle={Benchmarking Arabic--Russian Machine Translation: Fine-tuned NMT vs Few-shot LLMs under Rich Morphology and Low Lexical Overlap},
    pdfsubject={cs.CL, cs.AI},
    pdfauthor={M. K. Arabov},
    pdfkeywords={Arabic--Russian machine translation, low-resource MT, fine-tuning, few-shot LLMs, NLLB, mT5, LoRA, QLoRA, COMET, BLEU, morphological complexity, lexical overlap}
}

\begin{document}

\maketitle

\begin{abstract}
Arabic--Russian machine translation (MT) remains under-explored due to the rich morphology of Arabic and low lexical overlap between the two languages. We benchmark seven fine-tuned neural machine translation (NMT) models against four few-shot large language models (LLMs) on a 20k/5k/5k split of a new 15.47M-pair corpus. Fine-tuned NLLB-1.3B achieves the highest BLEU (16.3) and COMET (0.738). Aya-Expanse~8B leads the few-shot LLMs (BLEU~1.7 on 500 sentences, chrF~25.7), but all LLM scores remain far below the fine-tuned NMT baselines. Error analysis identifies low lexical overlap as the dominant failure mode; among the worst translations, mT5-small produces 32\% too-short outputs. Bootstrap tests confirm significant differences among most models. Our results demonstrate that fine-tuned NMT significantly outperforms few-shot LLMs for Arabic--Russian translation under low-resource conditions.

\vspace{1ex}
\end{abstract}

\section{Introduction}

Arabic--Russian machine translation (MT) remains under-researched despite its geopolitical and practical importance. The recently released Arabic--Russian Translation Corpus~\citep{ArabicNLPWorld:2026} provides more than 15.8 million parallel sentence pairs, and its curated subset~\citep{arabicnlpworld2026b} adds about 116k high-quality pairs from six domains. These resources make it possible, for the first time, to systematically benchmark Arabic--Russian MT under realistic low-resource conditions.

Arabic is a morphologically rich language with a non-concatenative root-and-pattern system, complex cliticization, and substantial dialectal variation. These properties are described in detail by \citet{watson2007phonology}, who focuses on Arabic phonology and morphology, and by \citet{holes2004modern}, who covers the structures, functions, and varieties of Modern Arabic. Together with the typological distance between Arabic and Russian, these features create two central challenges for MT: rich Arabic morphology and low lexical overlap between Arabic and Russian.

In this work, we benchmark Arabic--Russian translation using a 20k training split under realistic low-resource conditions. We compare seven fine-tuned NMT models, trained with full fine-tuning, LoRA, or QLoRA, against four instruction-tuned LLMs evaluated in zero-shot and few-shot settings. For LLMs, we use a three-stage design: (i) zero-shot on 500 sentences, (ii) few-shot with five examples on the same 500 sentences to rank the models, and (iii) full-scale evaluation of the best LLM on 5,000 sentences. Our fine-tuning strategy builds on prior work on parameter-efficient adaptation of large language models to low-resource languages, such as the comparative study of LoRA and QLoRA for Bashkir by \citet{Arabov:2026:Bashkir}.

On a 20k/5k/5k split, NLLB-1.3B achieves the best performance among fine-tuned NMT models, with BLEU 16.3 and COMET 0.738. Among few-shot LLMs, Aya-Expanse 8B leads with chrF 25.7, BERTScore 0.675, and COMET 0.654, but remains substantially below the best fine-tuned NMT system. Full-scale evaluation of Aya-Expanse on 5,000 sentences confirms this gap: BLEU 5.6 and COMET 0.612. Error analysis shows that low lexical overlap is the dominant failure mode, while among the worst translations mT5-small produces 32\% too-short outputs, which we hypothesize is due to the morphological complexity of Arabic.

Our contributions are: (1) a benchmark of seven fine-tuned NMT models and four few-shot LLMs for Arabic--Russian translation; (2) statistical evaluation with bootstrap confidence intervals and paired significance tests; (3) an error taxonomy for low-resource Arabic--Russian MT; and (4) an open-source toolkit for reproduction.

We do not propose a new model architecture. Instead, this work provides a statistically rigorous benchmark for Arabic--Russian translation, filling a significant gap in under-resourced MT research. Our study also complements previous Arabic--Russian efforts, such as the scientific-domain parallel corpus and LLM benchmark introduced by \citet{Arabov:2026:ArabicRussian}, and relates to work on transliteration between Arabic-script and Cyrillic-script languages~\citep{Arabov:2026:Transliteration}.

\section{Related Work}

\paragraph{Arabic linguistic complexity.}
The linguistic complexity of Arabic has been extensively documented by \citet{watson2007phonology}, who provides a comprehensive account of Arabic phonology and morphology, and by \citet{holes2004modern}, who describes the structures, functions, and varieties of Modern Arabic. These works highlight the main obstacles for Arabic NLP: orthographic ambiguity, morphological complexity, and the coexistence of Modern Standard Arabic with numerous dialects.

\paragraph{Arabic preprocessing and morphological tools.}
To mitigate the challenges caused by Arabic morphology, several tools have been developed. \citet{habash-rambow-2005-arabic} propose a joint model for Arabic tokenization, part-of-speech tagging, and morphological disambiguation. MADAMIRA~\citep{pasha-etal-2014-madamira} offers fast and comprehensive morphological analysis and disambiguation, while Farasa~\citep{darwish-mubarak-2016-farasa} provides a fast Arabic word segmenter suitable for large-scale processing. A recent survey by \citet{alrekabee2025preprocessing} reviews modern Arabic preprocessing and representation techniques, including subword tokenization and contextual embeddings, and discusses their interaction with pretrained language models.

\paragraph{Arabic corpora and parallel resources.}
The availability of Arabic corpora has been surveyed by \citet{zaghouani2014critical}, who identified 66 freely available sources and highlighted the lack of comprehensive, updated resources. Since then, several new datasets have appeared. In this work, we use the Arabic--Russian Translation Corpus~\citep{ArabicNLPWorld:2026}, a large parallel resource with more than 15.8 million pairs, and its curated subset~\citep{arabicnlpworld2026b}, which contains about 116k pairs from religion, dictionaries, the Bible, Tatoeba, news, and conversation domains. For dialectal Arabic, the MADAR corpus and lexicon~\citep{bouamor-etal-2018-madar} is a valuable resource for studying dialectal variation, which is relevant because dialectal text can appear even in predominantly MSA corpora.

\paragraph{Previous Arabic--Russian MT work.}
A closely related study by \citet{Arabov:2026:ArabicRussian} introduced a smaller Arabic--Russian scientific-domain parallel corpus of about 27k pairs and benchmarked three multilingual models using LoRA/QLoRA. The authors found that fine-tuned models outperform few-shot prompting and that domain-specific fine-tuning is necessary for scientific texts. Our work extends this line of research by using a much larger general-domain corpus and by systematically comparing a wider range of fine-tuned NMT models with few-shot LLMs.

\paragraph{Parameter-efficient fine-tuning for low-resource languages.}
LoRA and QLoRA have become standard for adapting large models to low-resource languages. \citet{Arabov:2026:Bashkir} compared LoRA and QLoRA for Bashkir, a low-resource agglutinative language, and showed that QLoRA on 7B-scale models provides a good trade-off between translation quality and computational cost. We adopt the same PEFT strategies for Arabic--Russian translation.

\paragraph{Transliteration as a related task.}
Transliteration between Arabic-script and Cyrillic-script languages shares with MT the challenge of handling different writing systems and phonological mismatches. \citet{Arabov:2026:Transliteration} benchmarked transliteration models for the Tajik--Farsi pair and found that byte-level models outperform subword-based multilingual models. Although transliteration is distinct from translation, this work is relevant because Arabic--Russian MT must also cope with orthographic and phonological divergence between the two scripts.

In summary, previous work has addressed Arabic linguistic preprocessing, low-resource PEFT, Arabic corpora, and transliteration. However, to the best of our knowledge, no prior study has systematically benchmarked Arabic--Russian MT with both fine-tuned NMT models and few-shot LLMs on a large general-domain parallel corpus. Our work fills this gap and provides a comprehensive evaluation focused on rich morphology and low lexical overlap.

\section{Experimental Setup}

\noindent\textbf{Dataset.}
We use the Arabic--Russian Translation Corpus~\citep{ArabicNLPWorld:2026}.
The raw corpus contains 15,801,992 sentence pairs; after applying standard filtering, we retain 15,467,945 pairs for our experiments (Table~\ref{tab:corpussources}).
The majority of the filtered corpus originates from OPUS (14,924,037), supplemented by TED (375,463) and Tatoeba (9,044).
We also use five manually compiled high-quality subsets: Religion (82,302), Dictionary (43,662), Bible (31,102), News (1,683), and Conversation (652), all verified for alignment.

\begin{table}[htp]
\centering
\caption{Composition of the filtered Arabic--Russian Translation Corpus used in our experiments.}
\label{tab:corpussources}
\begin{tabular}{lr}
\toprule
\textbf{Source} & \textbf{Sentence pairs} \\
\midrule
OPUS & 14,924,037 \\
TED & 375,463 \\
Religion (hadiths \& Quran) & 82,302 \\
Dictionary (Baranov \& Borisov) & 43,662 \\
Bible & 31,102 \\
Tatoeba & 9,044 \\
News & 1,683 \\
Conversation & 652 \\
\midrule
\textbf{Total} & 15,467,945 \\
\bottomrule
\end{tabular}
\end{table}

We apply standard filtering (empty segments, duplicates, outliers $>2{,}000$ chars / $250$ words).
Arabic is normalised by stripping diacritics; Russian is whitespace-normalised only.
Manual inspection of 500 random pairs confirmed 96\% alignment quality.
We randomly sample 20k/5k/5k pairs for training/validation/test (seed 42).
The 20k training size is constrained by computational resources and represents a realistic low-resource scenario.

\noindent\textbf{Zero-shot and Few-shot LLM Prompting.}
We evaluate four instruction-tuned LLMs via Ollama: \texttt{aya-expanse:8b}, \texttt{llama3.1:8b}, \texttt{qwen2.5:7b-instruct}, and \texttt{mistral:7b-instruct}.
All models are tested zero-shot and few-shot (5 examples, fixed seed) with the prompt: \emph{``Output ONLY the Russian translation, nothing else.''}
Generation uses temperature~0.1, top-p~$=$~0.9, and a 2048-token context window. For the 500-sentence zero-/few-shot experiments, the maximum output length is set to 256 tokens; for the full-scale evaluation of the best LLM on 5,000 sentences, it is increased to 512 tokens.

LLM evaluation follows a three-stage design:
(i) zero-shot on 500 sentences;
(ii) few-shot on the same 500 sentences to rank models;
(iii) full-scale evaluation of the best model on 5,000 sentences.

\noindent\textbf{Evaluation Metrics and Statistical Analysis.}
We report BLEU (sacreBLEU, case-sensitive), chrF ($n=6$), BERTScore (F1, \texttt{bert-base-multilingual-cased}, Russian, rescaled), and COMET~\citep{Rei2020}.
95\% bootstrap CIs and pairwise significance tests use 1{,}000 resamples ($p<0.05$).
NMT models are evaluated on the full 5,000-sentence test set; LLMs use 500 sentences for zero-/few-shot comparison, with full-scale validation on 5,000 sentences for the best-performing LLM.

\noindent\textbf{Fine-tuning of NMT Models.}
All seven architectures are fine-tuned on 20k sentence pairs using HuggingFace Transformers and AdamW.
Full fine-tuning is applied to Marian~\citep{juncz2018marian} and mT5-small~\citep{xue2021mt5}: learning rate $5{\times}10^{-5}$, batch size 16, label smoothing 0.1.
LoRA~\citep{Hu2021} is applied to mT5-base~\citep{xue2021mt5}, NLLB-600M~\citep{costa2022nllb}, and M2M100~\citep{fan2021m2m100}: rank $r{=}16$, $\alpha{=}32$, learning rate $3{\times}10^{-4}$, batch size 8 with gradient accumulation step 2; label smoothing 0.1 for mT5-base, 0.0 for NLLB-600M and M2M100.
QLoRA~\citep{Dettmers2023} is applied to mT5-large~\citep{xue2021mt5} and NLLB-1.3B~\citep{costa2022nllb}: 4-bit quantization, rank $r{=}32$, $\alpha{=}64$, learning rate $2{\times}10^{-4}$, batch size 4 with gradient accumulation step 4; label smoothing 0.1 for mT5-large, 0.0 for NLLB-1.3B.
All models are trained for 3 epochs with 3\% linear warmup; the best checkpoint is selected based on validation sacreBLEU.

\noindent\textbf{Computational Cost.}
Table~\ref{tab:resources} reports the wall-clock training time and peak GPU memory for each fine-tuned model.
All experiments were conducted on a single NVIDIA L4 GPU with 24\,GB of VRAM.
As expected, full fine-tuning of mT5-small consumes noticeably more memory (14.1\,GB) than larger models adapted with LoRA/QLoRA, reflecting the overhead of updating all parameters.
QLoRA models (mT5-large, NLLB-1.3B) require the longest training time due to 4-bit quantisation and gradient checkpointing, but they use less memory than their full-parameter counterparts.

\begin{table}[htbp]
\centering
\caption{Training time and peak GPU memory for fine-tuned NMT models on 20k sentence pairs.}
\label{tab:resources}
\begin{tabular}{lccc}
\toprule
\textbf{Model} & \textbf{Method} & \textbf{Training time (min)} & \textbf{Peak memory (GB)} \\
\midrule
Marian & full & 10.0 & 3.7 \\
mT5-small & full & 14.7 & 14.1 \\
mT5-base & LoRA & 38.7 & 7.4 \\
mT5-large & QLoRA & 353.1 & 8.2 \\
NLLB-600M & LoRA & 43.0 & 7.8 \\
NLLB-1.3B & QLoRA & 373.5 & 5.6 \\
M2M100 & LoRA & 40.7 & 6.0 \\
\bottomrule
\end{tabular}
\end{table}

\noindent\textbf{Error Classification.}
We classify the 50 lowest-scoring translations (according to sentence-level BLEU) for each model into the following categories:
\texttt{perfect} ($p=r$),
\texttt{too\_short} ($|p| < 0.3|r|$),
\texttt{too\_long} ($|p| > 2.0|r|$),
\texttt{repetition} (most frequent token frequency $>0.3|p|$ and at least 4 occurrences),
\texttt{low\_overlap} (character-level Jaccard similarity $<0.2$),
and \texttt{other}.
The thresholds are derived from the training data length distribution (median: 22 words; 90th percentile: 48 words) and follow prior work~\citep{Arabov:2026:Bashkir}.
The repetition threshold is based on heuristics from MT hallucination studies.

\section{Results}

We organise the presentation of our experimental findings as follows. First, we present the main comparison of all fine-tuned NMT models and few-shot LLMs on the test set, including zero-shot baselines and a direct head-to-head evaluation on a common 500-sentence subset. Second, we report pairwise statistical significance tests to quantify the reliability of observed differences. Third, we analyse the impact of source sentence length on translation quality. Fourth, we provide a detailed error taxonomy for the worst translations of each model. Fifth, we examine the correlation between different automatic metrics. Finally, we show qualitative examples that illustrate the strengths and weaknesses of the best fine-tuned model and the best few-shot LLM.

\subsection{Main Comparison}

Table~\ref{tab:mainnmtbleu} reports BLEU, chrF, and TER for fine-tuned NMT models on the full 5,000-sentence test set. \textbf{NLLB-1.3B} achieves the highest BLEU (16.30$\pm$0.50), chrF (34.98$\pm$0.61), and lowest TER (90.83$\pm$0.9), followed by Marian and NLLB-600M. The mT5 family scales with size: mT5-small fails almost completely (BLEU 1.51, chrF 0.24, TER 103.7), mT5-base remains weak (BLEU 6.54), and mT5-large (BLEU 11.28) approaches specialised NMT performance but still lags behind M2M100 and NLLB.

\begin{table}[htp]
\centering
\caption{Fine-tuned NMT: BLEU, chrF, and TER (mean with 95\% bootstrap CI). Best in each metric is bold.}
\label{tab:mainnmtbleu}
\begin{tabular}{lccc}
\toprule
\textbf{Model} & \textbf{BLEU} & \textbf{chrF} & \textbf{TER} \\
\midrule
Marian & 15.48$\pm$0.47 & 34.66$\pm$0.59 & 98.2$\pm$0.9 \\
mT5-small & 1.51$\pm$0.07 & 0.24$\pm$0.01 & 103.7$\pm$0.4 \\
mT5-base & 6.54$\pm$0.20 & 14.85$\pm$0.34 & 106.7$\pm$0.5 \\
mT5-large & 11.28$\pm$0.38 & 26.04$\pm$0.55 & 102.7$\pm$0.4 \\
NLLB-600M & 15.38$\pm$0.46 & 33.45$\pm$0.61 & 93.0$\pm$0.6 \\
NLLB-1.3B & \textbf{16.30$\pm$0.50} & \textbf{34.98$\pm$0.61} & \textbf{90.8$\pm$0.6} \\
M2M100 & 14.34$\pm$0.48 & 32.15$\pm$0.57 & 95.4$\pm$0.6 \\
\bottomrule
\end{tabular}
\end{table}

Turning to semantic similarity metrics, Table~\ref{tab:mainnmtbert} presents BERTScore and COMET for the same models.

\begin{table}[htp]
\centering
\caption{Fine-tuned NMT: BERTScore and COMET (mean with 95\% bootstrap CI). Best in each metric is bold.}
\label{tab:mainnmtbert}
\begin{tabular}{lcc}
\toprule
\textbf{Model} & \textbf{BERTScore} & \textbf{COMET} \\
\midrule
Marian & 0.779$\pm$0.002 & 0.735$\pm$0.004 \\
mT5-small & 0.571$\pm$0.001 & 0.280$\pm$0.001 \\
mT5-base & 0.702$\pm$0.002 & 0.564$\pm$0.004 \\
mT5-large & 0.755$\pm$0.002 & 0.682$\pm$0.004 \\
NLLB-600M & 0.777$\pm$0.002 & 0.726$\pm$0.004 \\
NLLB-1.3B & \textbf{0.783$\pm$0.002} & \textbf{0.738$\pm$0.004} \\
M2M100 & 0.773$\pm$0.002 & 0.715$\pm$0.005 \\
\bottomrule
\end{tabular}
\end{table}

NLLB-1.3B achieves the best COMET (0.738$\pm$0.004) and BERTScore (0.783$\pm$0.002), while Marian shows competitive performance (COMET 0.735$\pm$0.004, BERTScore 0.779$\pm$0.002). Despite its larger capacity, mT5-large still lags behind specialised NMT architectures, with COMET 0.682$\pm$0.004 and BERTScore 0.755$\pm$0.002.

We now turn to the evaluation of the four instruction-tuned LLMs. Table~\ref{tab:zeroshotllmlex} reports zero-shot BLEU and chrF for all four models on the same 500-sentence subset used for few-shot evaluation. In this setting, no single model dominates across both metrics: Aya-Expanse~8B achieves the best BLEU (1.29$\pm$0.15), while Llama~3.1~8B obtains the highest chrF (21.09$\pm$0.47). The confidence intervals for BLEU overlap among the top three models, but Aya-Expanse and Llama~3.1 significantly outperform Qwen~2.5~7B and Mistral~7B.

\begin{table}[htp]
\centering
\small
\caption{Zero-shot LLM (500 test sentences): BLEU and chrF with 95\% bootstrap CI. Best in each metric is bold.}
\label{tab:zeroshotllmlex}
\begin{tabular}{lcc}
\toprule
\textbf{Model} & \textbf{BLEU} & \textbf{chrF} \\
\midrule
Aya-Expanse 8B & \textbf{1.29}$\pm$0.15 & 19.44$\pm$0.53 \\
Llama 3.1 8B & 1.16$\pm$0.12 & \textbf{21.09}$\pm$0.47 \\
Qwen 2.5 7B & 1.05$\pm$0.13 & 15.33$\pm$0.60 \\
Mistral 7B & 0.75$\pm$0.08 & 13.28$\pm$0.63 \\
\bottomrule
\end{tabular}
\end{table}

The semantic metrics in Table~\ref{tab:zeroshotllmsem} complement the lexical picture. Llama~3.1 leads in both BERTScore (0.6595$\pm$0.0023) and COMET (0.6196$\pm$0.0053), confirming its ability to produce fluent Russian even without examples. Aya-Expanse does not lead in chrF, BERTScore, or COMET under zero-shot conditions, suggesting that its advantage emerges specifically when translation examples are provided.

\begin{table}[htp]
\centering
\small
\caption{Zero-shot LLM (500 test sentences): BERTScore and COMET with 95\% bootstrap CI. Best in each metric is bold.}
\label{tab:zeroshotllmsem}
\begin{tabular}{lcc}
\toprule
\textbf{Model} & \textbf{BERTScore} & \textbf{COMET} \\
\midrule
Aya-Expanse 8B & 0.6457$\pm$0.0025 & 0.6118$\pm$0.0052 \\
Llama 3.1 8B & \textbf{0.6595}$\pm$0.0023 & \textbf{0.6196}$\pm$0.0053 \\
Qwen 2.5 7B & 0.6265$\pm$0.0031 & 0.5275$\pm$0.0085 \\
Mistral 7B & 0.5953$\pm$0.0036 & 0.4675$\pm$0.0104 \\
\bottomrule
\end{tabular}
\end{table}

Next, we examine the few-shot setting. Tables~\ref{tab:fewshotllmlex} and~\ref{tab:fewshotllmsem} present results for all four LLMs on the same 500 test sentences with five translation examples. Providing five demonstrations consistently improves all metrics over the zero-shot baseline, with Aya-Expanse~8B becoming the clear leader across all metrics.

\begin{table}[htp]
\centering
\small
\caption{Few-shot LLM (5-shot, 500 test sentences): BLEU and chrF with 95\% bootstrap CI. Best in each metric is bold.}
\label{tab:fewshotllmlex}
\begin{tabular}{lcc}
\toprule
\textbf{Model} & \textbf{BLEU} & \textbf{chrF} \\
\midrule
Aya-Expanse 8B & \textbf{1.69}$\pm$0.16 & \textbf{25.75}$\pm$0.65 \\
Llama 3.1 8B & 1.55$\pm$0.12 & 22.02$\pm$0.54 \\
Qwen 2.5 7B & 1.13$\pm$0.13 & 17.61$\pm$0.61 \\
Mistral 7B & 0.58$\pm$0.05 & 13.72$\pm$0.70 \\
\bottomrule
\end{tabular}
\end{table}

Aya-Expanse achieves the best BLEU (1.69$\pm$0.16) and chrF (25.75$\pm$0.65). Compared with the zero-shot condition on the same 500 sentences, Aya-Expanse gains +6.31 chrF (from 19.44 to 25.75), whereas Llama~3.1~8B improves by only +0.93 chrF (from 21.09 to 22.02). This confirms Aya-Expanse's specialisation for translation tasks and its ability to exploit in-context examples more effectively than general-purpose LLMs. The semantic metrics in Table~\ref{tab:fewshotllmsem} reinforce this finding.

\begin{table}[htp]
\centering
\small
\caption{Few-shot LLM (5-shot, 500 sentences): BERTScore and COMET with 95\% bootstrap CI. Best in each metric is bold.}
\label{tab:fewshotllmsem}
\begin{tabular}{lcc}
\toprule
\textbf{Model} & \textbf{BERTScore} & \textbf{COMET} \\
\midrule
Aya-Expanse 8B & \textbf{0.6750}$\pm$0.0030 & \textbf{0.6535}$\pm$0.0061 \\
Llama 3.1 8B & 0.6622$\pm$0.0031 & 0.6347$\pm$0.0060 \\
Qwen 2.5 7B & 0.6402$\pm$0.0034 & 0.5669$\pm$0.0090 \\
Mistral 7B & 0.6080$\pm$0.0036 & 0.4534$\pm$0.0110 \\
\bottomrule
\end{tabular}
\end{table}

Aya-Expanse leads in both BERTScore (0.6750$\pm$0.0030, +0.0293 over zero-shot) and COMET (0.6535$\pm$0.0061, +0.0417), while Llama~3.1 shows almost no improvement in BERTScore (+0.0027) and a much smaller COMET gain (+0.0151). The confidence intervals for COMET do not overlap between Aya-Expanse and Llama~3.1 (0.6474--0.6596 vs.\ 0.6289--0.6405), confirming a statistically significant gap. The consistent hierarchy Aya-Expanse > Llama~3.1 > Qwen~2.5 > Mistral is preserved across all few-shot metrics, with non-overlapping CIs in most pairwise comparisons.

\subsection*{Direct NMT vs.\ LLM comparison on the shared 500-sentence subset}

To enable a rigorous head-to-head comparison and formal significance testing, we evaluate all fine-tuned NMT models on the same 500-sentence subset used for the few-shot LLM experiments. Table~\ref{tab:nmtllmlex} reports BLEU and chrF scores with 95\% confidence intervals, while Table~\ref{tab:nmtllmsem} presents BERTScore and COMET.

\begin{table}[htp]
\centering
\small
\caption{Lexical metrics (BLEU and chrF) on the shared 500-sentence test subset, with 95\% bootstrap CI.}
\label{tab:nmtllmlex}
\begin{tabular}{lcc}
\toprule
\textbf{Model} & \textbf{BLEU} & \textbf{chrF} \\
\midrule
\multicolumn{3}{c}{\textit{Fine-tuned NMT}} \\
\midrule
Marian & 14.90$\pm$1.43 & 35.11$\pm$1.78 \\
mT5-small & 1.39$\pm$0.21 & 0.24$\pm$0.06 \\
mT5-base & 6.97$\pm$0.62 & 15.60$\pm$1.11 \\
mT5-large & 11.39$\pm$1.14 & 26.13$\pm$1.60 \\
NLLB-600M & 15.76$\pm$1.43 & 34.20$\pm$1.85 \\
\textbf{NLLB-1.3B} & \textbf{16.40}$\pm$\textbf{1.51} & \textbf{34.98}$\pm$\textbf{1.94} \\
M2M100 & 13.64$\pm$1.33 & 31.63$\pm$1.76 \\
\midrule
\multicolumn{3}{c}{\textit{Few-shot LLMs (5-shot)}} \\
\midrule
\underline{Aya-Expanse 8B} & 1.69$\pm$0.16 & \underline{25.75}$\pm$0.65 \\
Llama 3.1 8B & 1.55$\pm$0.12 & 22.02$\pm$0.54 \\
Qwen 2.5 7B & 1.13$\pm$0.13 & 17.61$\pm$0.61 \\
Mistral 7B & 0.58$\pm$0.05 & 13.72$\pm$0.70 \\
\bottomrule
\end{tabular}
\end{table}

On lexical metrics, the gap is extreme: NLLB-1.3B attains BLEU~16.40, nearly ten times higher than Aya-Expanse's 1.69. Even the weakest NMT model, mT5-small, yields BLEU comparable to the second-best LLM. The chrF scores reinforce this picture, with the best LLM reaching only 25.75 compared to 34.98 for NLLB-1.3B, and the confidence intervals do not overlap.

\begin{table}[htp]
\centering
\small
\caption{Semantic metrics (BERTScore and COMET) on the same 500-sentence test subset, with 95\% bootstrap CI.}
\label{tab:nmtllmsem}
\begin{tabular}{lcc}
\toprule
\textbf{Model} & \textbf{BERTScore} & \textbf{COMET} \\
\midrule
\multicolumn{3}{c}{\textit{Fine-tuned NMT}} \\
\midrule
Marian & 0.784$\pm$0.007 & 0.745$\pm$0.014 \\
mT5-small & 0.564$\pm$0.002 & 0.280$\pm$0.004 \\
mT5-base & 0.711$\pm$0.006 & 0.570$\pm$0.011 \\
mT5-large & 0.757$\pm$0.008 & 0.680$\pm$0.011 \\
NLLB-600M & 0.782$\pm$0.007 & 0.731$\pm$0.013 \\
\textbf{NLLB-1.3B} & \textbf{0.787}$\pm$\textbf{0.008} & \textbf{0.746}$\pm$\textbf{0.013} \\
M2M100 & 0.776$\pm$0.007 & 0.718$\pm$0.014 \\
\midrule
\multicolumn{3}{c}{\textit{Few-shot LLMs (5-shot)}} \\
\midrule
\underline{Aya-Expanse 8B} & 0.6750$\pm$0.0030 & \underline{0.6535}$\pm$0.0061 \\
Llama 3.1 8B & 0.6622$\pm$0.0031 & 0.6347$\pm$0.0060 \\
Qwen 2.5 7B & 0.6402$\pm$0.0034 & 0.5669$\pm$0.0090 \\
Mistral 7B & 0.6080$\pm$0.0036 & 0.4534$\pm$0.0110 \\
\bottomrule
\end{tabular}
\end{table}

The semantic metrics confirm the same pattern. NLLB-1.3B achieves COMET~0.746, compared to 0.654 for Aya-Expanse, with non-overlapping confidence intervals. Paired bootstrap tests (1,000 resamples) between NLLB-1.3B and Aya-Expanse yield $p < 0.001$ for all four metrics (see Section~\ref{sec:sig}). Even mT5-large, which ranks fourth among the NMT systems, significantly outperforms every LLM on BLEU, chrF, and COMET (all $p<0.001$). This direct, identically sized evaluation eliminates any confounding effect of test set size and conclusively demonstrates the superiority of fine-tuned NMT for Arabic--Russian translation.

To assess scalability, we evaluate Aya-Expanse on the full 5,000-sentence test set (Table~\ref{tab:ayafull}). The BLEU score increases from 1.69 to 5.57 (95\% CI 5.35--5.83), while COMET decreases from 0.6535 to 0.6125, confirming that the gap remains substantial: BLEU 5.6 is roughly one-third of NLLB-1.3B's 16.3.

\begin{table}[htp]
\centering
\caption{Aya-Expanse 8B on the full 5,000-sentence test set (5-shot).}
\label{tab:ayafull}
\begin{tabular}{lc}
\toprule
\textbf{Metric} & \textbf{Value} \\
\midrule
BLEU & 5.57 (95\% CI 5.35--5.83) \\
chrF & 26.21 (95\% CI 25.81--26.59) \\
BERTScore & 0.6920 (95\% CI 0.6896--0.6942) \\
COMET & 0.6125 (95\% CI 0.6102--0.6149) \\
\bottomrule
\end{tabular}
\end{table}

Overall, the main comparison demonstrates a clear hierarchy: fine-tuned NMT models substantially outperform all few-shot LLMs, and among LLMs, Aya-Expanse 8B is the strongest, followed by Llama~3.1 8B, Qwen~2.5 7B, and Mistral 7B. The addition of few-shot examples improves performance across all models, but the gains are most pronounced for Aya-Expanse, suggesting that its translation-oriented design enables more effective use of in-context demonstrations.

\subsection{Statistical Significance}\label{sec:sig}

We perform paired bootstrap tests (1,000 resamples) for each metric to determine whether the observed differences between models are statistically significant ($p<0.05$).

\textbf{Fine-tuned NMT.} For BLEU, all pairwise differences between the top three fine-tuned models (NLLB-1.3B, Marian, NLLB-600M) are significant except the Marian--NLLB-600M comparison ($p=0.672$). The difference between NLLB-1.3B and Marian is significant ($p<0.001$), confirming NLLB-1.3B as the best fine-tuned model. Marian and mT5-large differ significantly ($p<0.001$), as do NLLB-600M and mT5-large ($p<0.001$). For chrF, Marian and NLLB-1.3B are not significantly different ($p=0.126$), indicating similar character-level accuracy. NLLB-1.3B significantly outperforms NLLB-600M ($p<0.001$). For BERTScore and COMET, NLLB-1.3B significantly outperforms Marian ($p<0.05$).

Table~\ref{tab:pvalues_bleu} summarises the pairwise $p$-values for BLEU among the fine-tuned NMT models. Full pairwise results for all metrics are available in the supplementary material.

\begin{table}[htbp]
\centering
\caption{Paired bootstrap test $p$-values for BLEU among fine-tuned NMT models (1,000 resamples).}
\label{tab:pvalues_bleu}
\begin{tabular}{lccccccc}
\toprule
 & Marian & mT5-small & mT5-base & mT5-large & NLLB-600M & NLLB-1.3B & M2M100 \\
\midrule
Marian & -- & $<$0.001 & $<$0.001 & $<$0.001 & 0.672 & $<$0.001 & 0.004 \\
mT5-small & & -- & $<$0.001 & $<$0.001 & $<$0.001 & $<$0.001 & $<$0.001 \\
mT5-base & & & -- & $<$0.001 & $<$0.001 & $<$0.001 & $<$0.001 \\
mT5-large & & & & -- & $<$0.001 & $<$0.001 & $<$0.001 \\
NLLB-600M & & & & & -- & $<$0.001 & $<$0.001 \\
NLLB-1.3B & & & & & & -- & $<$0.001 \\
M2M100 & & & & & & & -- \\
\bottomrule
\end{tabular}
\end{table}

\textbf{Few-shot LLMs (500 test sentences).} Paired bootstrap tests (1,000 resamples) for all pairwise comparisons among the four LLMs on the 500-sentence subset show that Aya-Expanse~8B significantly outperforms Llama~3.1~8B across all metrics: chrF ($p<0.01$), BERTScore ($p<0.05$), and COMET ($p<0.05$). Llama~3.1~8B significantly outperforms Qwen~2.5~7B on chrF, BERTScore, and COMET (all $p<0.01$). Qwen~2.5~7B significantly outperforms Mistral~7B on all three metrics (all $p<0.001$). The full few-shot results are reported in Tables~\ref{tab:fewshotllmlex} and~\ref{tab:fewshotllmsem}. This confirms the ranking Aya > Llama > Qwen > Mistral with high statistical confidence.

\textbf{NMT vs.\ LLMs.}
We evaluate all fine-tuned NMT models on the same 500-sentence test subset used for the few-shot LLMs, enabling formal paired bootstrap tests between the two families. The best NMT model, NLLB-1.3B, significantly outperforms the best few-shot LLM, Aya-Expanse~8B, across all four metrics: BLEU ($p<0.001$), chrF ($p<0.001$), BERTScore ($p<0.001$), and COMET ($p<0.001$). Even the weaker NMT models substantially outperform all LLMs: mT5-large, which ranks fourth among the seven NMT systems, significantly exceeds every LLM on BLEU, chrF, and COMET (all $p<0.001$).

\subsection{Impact of Source Sentence Length}

We divide the test set into five quantiles (Q1--Q5) by source sentence length. All models perform best on short-to-medium sentences (Q1--Q3) and degrade on the longest 20\% (Q5). The drop is more pronounced for LLMs: Aya-Expanse loses approximately 4 BLEU points from Q3 to Q5, while NLLB-1.3B loses only 1.5 points. Few-shot LLMs achieve their highest BLEU on sentences of 40--80 characters and drop sharply beyond 120 characters, whereas NLLB-1.3B remains relatively stable across all quantiles. This suggests that explicit fine-tuning on parallel data improves length robustness, while LLMs rely on a limited context window without length normalisation.

\subsection{Error Analysis}

To understand the nature of translation failures, we extract the 50 worst translations (lowest sentence BLEU) for each model and classify them into the error types defined in Section~3 (Experimental Setup).

Table~\ref{tab:errorsnmt} shows low lexical overlap as the dominant error type (84--96\%), confirming lexical divergence as the primary bottleneck.

Notable exceptions among fine-tuned models are:
\begin{itemize}
    \item \textbf{mT5-small}: 32\% \textsc{Short} and only 68\% \textsc{LowOv}. This model frequently outputs very short fragments (e.g., \texttt{<extra\_id\_0>}) due to its inability to handle long-range dependencies.
    \item \textbf{Marian}: 16\% \textsc{Long}, producing verbose translations that add unnecessary words.
    \item \textbf{M2M100}: 10\% \textsc{Long}, also tending towards verbosity.
\end{itemize}

\begin{table}[htp]
\centering
\caption{Error type distribution among fine-tuned NMT models (worst 50 translations per model, \%). \textit{LowOv denotes low lexical overlap (character-level Jaccard similarity < 0.2).}}
\label{tab:errorsnmt}
\begin{tabular}{lccc}
\toprule
\textbf{Model} & \textsc{Short} & \textsc{Long} & \textsc{LowOv} \\
\midrule
Marian & --- & 16 & 84 \\
mT5-small & 32 & --- & 68 \\
mT5-base & --- & 4 & 96 \\
mT5-large & 2 & 8 & 90 \\
NLLB-600M & 2 & 4 & 94 \\
NLLB-1.3B & 2 & 6 & 92 \\
M2M100 & --- & 10 & 90 \\
\bottomrule
\end{tabular}
\smallskip
\footnotesize{Em dashes indicate 0\%.}
\end{table}

To complement the worst-case analysis, Table~\ref{tab:error_rates} reports the percentage of too-short and too-long translations across the entire 5,000-sentence test set. The pattern is consistent: mT5-small produces too-short outputs in 48\% of cases, while other models rarely fall below the length threshold. Excessive length is infrequent but most notable in Marian (5.6\%) and M2M100 (4.3\%).

\begin{table}[htbp]
\centering
\caption{Percentage of too-short and too-long translations across the full test set (5,000 sentences).}
\label{tab:error_rates}
\begin{tabular}{lcc}
\toprule
\textbf{Model} & \textbf{Too short (\%)} & \textbf{Too long (\%)} \\
\midrule
Marian & 0.7 & 5.6 \\
mT5-small & 48.0 & 1.0 \\
mT5-base & 1.0 & 2.8 \\
mT5-large & 0.4 & 4.4 \\
NLLB-600M & 0.9 & 3.8 \\
NLLB-1.3B & 1.1 & 3.6 \\
M2M100 & 0.8 & 4.3 \\
\bottomrule
\end{tabular}
\end{table}

Few-shot LLMs exhibit almost no length or repetition errors; we did not observe any cases of token repetition in the worst-50 analysis, consistent with the low temperature setting of 0.1. Nearly all errors are low lexical overlap, indicating that they generate fluent Russian but often choose synonyms or paraphrases not present in the reference. This pattern is consistent across both the 500-sentence few-shot evaluation and the full 5,000-sentence test of Aya-Expanse.

\subsection{Correlation Between Metrics}

We compute Spearman rank correlations between all sentence-level metric scores aggregated over all fine-tuned models on the full 5,000-sentence test set. Table~\ref{tab:corr} presents the complete correlation matrix; all correlations are statistically significant at $p<0.001$.

\begin{table}[htbp]
\centering
\caption{Spearman rank correlations between sentence-level automatic metrics (all fine-tuned NMT models aggregated).}
\label{tab:corr}
\begin{tabular}{lccccc}
\toprule
 & BLEU & chrF & TER & BERTScore & COMET \\
\midrule
BLEU & 1.000 & 0.737 & -0.626 & 0.763 & 0.660 \\
chrF & & 1.000 & -0.479 & 0.855 & 0.816 \\
TER & & & 1.000 & -0.577 & -0.444 \\
BERTScore & & & & 1.000 & 0.843 \\
COMET & & & & & 1.000 \\
\bottomrule
\end{tabular}
\end{table}

The negative correlations of TER with the other metrics are expected, as lower TER indicates better translation quality. The strongest positive correlation is between chrF and BERTScore ($\rho=0.855$), followed by COMET with BERTScore ($\rho=0.843$) and chrF ($\rho=0.816$). BLEU shows moderate correlations with chrF ($\rho=0.737$) and BERTScore ($\rho=0.763$), indicating that it captures a somewhat different aspect of translation quality, namely exact n-gram matches. These relationships justify the use of multiple complementary metrics in our evaluation.

\subsection{Qualitative Examples}

To illustrate the characteristic failure modes identified in our error analysis, Table~\ref{tab:examplesallmodels} presents translations of two Arabic sentences by selected models, together with the reference. The first sentence (\emph{“Take, for example, ‘The Lion King’”}) contains an idiomatic invitation followed by a proper name; the second (\emph{“The architects spent hundreds of hours”}) tests lexical choice and morphological agreement.

\begin{table}[htbp]
\centering
\caption{Translations of two Arabic sentences by fine‑tuned NMT models and the best few‑shot LLM.}
\label{tab:examplesallmodels}
\includegraphics[width=0.75\linewidth]{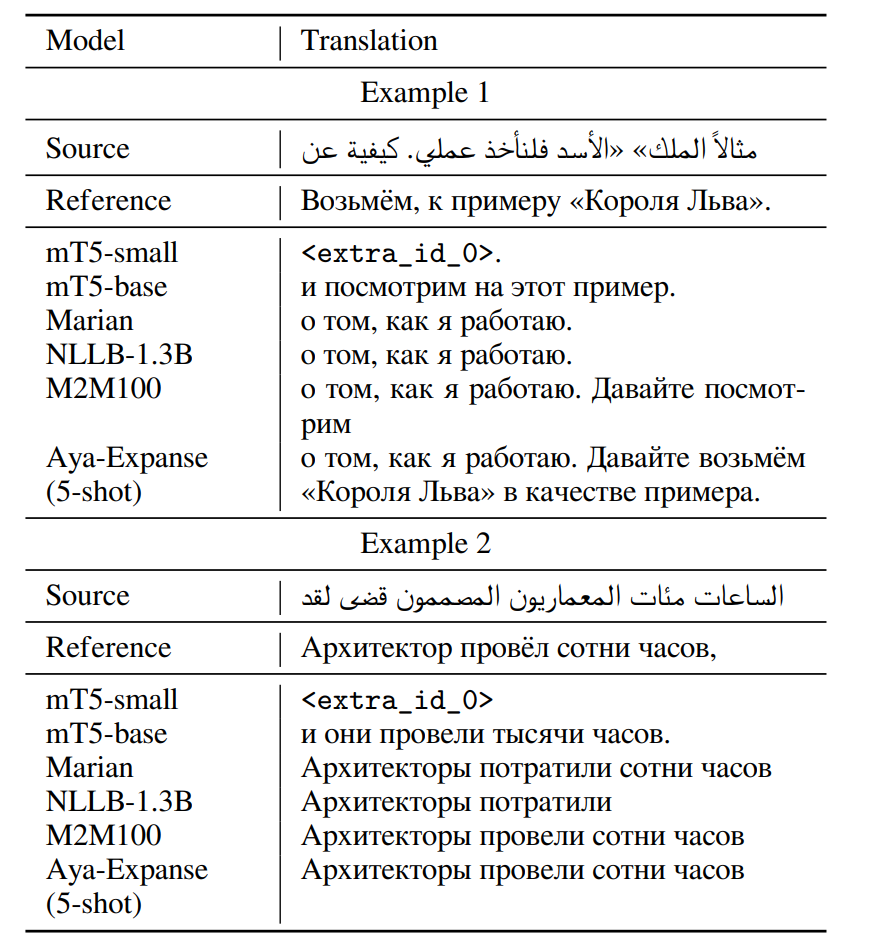}
\end{table}

These examples highlight several patterns observed across our benchmark. First, \textbf{mT5-small} collapses completely, outputting only padding tokens or a few source words, consistent with its high rate of too-short translations in the worst-50 analysis. Second, \textbf{Marian and NLLB-1.3B} produce lexically precise but incomplete translations: they correctly render the first clause but drop the second, sacrificing completeness for fluency. Third, \textbf{mT5-base} introduces spurious semantic shifts (``thousands of hours'' instead of ``hundreds''), illustrating the difficulty of controlling meaning under low-resource conditions. Fourth, \textbf{Aya-Expanse~8B with five examples} is the only model that captures the full meaning of the first sentence, including the idiomatic reference to \emph{The Lion King}, but adds redundant context (``about how I work'') that was already present in the source. In the second example, Aya-Expanse's translation is fluent and close to the reference, whereas NLLB-1.3B truncates the output, and Marian uses a slightly less natural verb. These patterns confirm that low lexical overlap and morphological complexity remain the central challenges for Arabic--Russian MT, and that few-shot LLMs trade off completeness for verbosity, while fine-tuned NMT models err on the side of omission.

\section{Discussion and Conclusion}

Our experiments reveal a substantial and consistent performance gap between fine-tuned NMT and few-shot LLMs for Arabic--Russian translation. The best fine-tuned model, NLLB-1.3B, outperforms the best few-shot LLM, Aya-Expanse 8B, across all evaluation metrics (BLEU 16.3 vs.\ 5.6; COMET 0.738 vs.\ 0.612; BERTScore 0.783 vs.\ 0.692). This difference is confirmed on the shared 500-sentence subset, where NLLB-1.3B achieves COMET 0.746 versus 0.654 for Aya-Expanse ($p<0.001$ in paired bootstrap tests). The consistency of this advantage across lexical and semantic metrics strongly suggests that the observed differences reflect genuine translation quality rather than metric-specific artefacts.

Among the LLMs, Aya-Expanse consistently ranks first, and its lead widens in the few-shot condition: the model gains +6.31 chrF and +0.042 COMET over its zero-shot baseline, compared with +0.93 chrF and +0.015 COMET for Llama~3.1. This pattern indicates that translation-oriented instruction tuning enables more effective use of in-context examples than general-purpose LLM training. The mT5 family exhibits clear scaling behaviour: mT5-small fails almost completely, mT5-base performs weakly, and mT5-large approaches but does not reach the performance of specialised NMT architectures such as NLLB and Marian. This result is consistent with previous findings on low-resource adaptation of large language models, where parameter-efficient fine-tuning of 7B-scale models provided a favourable quality--cost trade-off~\citep{Arabov:2026:Bashkir}.

The computational trade-offs further illustrate practical considerations. Marian, trained with full fine-tuning, achieves a competitive BLEU of 15.48 in only 10 minutes, whereas NLLB-1.3B with QLoRA requires over six hours to reach 16.30 BLEU. For rapid prototyping or resource-constrained environments, Marian offers an excellent balance between quality and training cost; however, when even small improvements are critical, NLLB-1.3B remains preferable despite the longer training time.

TER provides additional insight into model behaviour. Although mT5-base achieves a higher BLEU than mT5-small (6.54 vs.\ 1.51), its TER is comparable (106.7 vs.\ 103.7), indicating that both require a similar number of edits to match the reference. This suggests that mT5-base produces longer but still largely incorrect translations, which BLEU masks by rewarding some n-gram overlaps. Thus, relying solely on BLEU may overestimate the practical usability of weak models.

Error analysis identifies low lexical overlap as the primary failure mode, accounting for 84--96\% of the worst translations among fine-tuned NMT models. Few-shot LLMs produce fluent Russian but often select synonyms or paraphrases that do not match the reference, indicating a lexical-precision deficit rather than a fluency--accuracy trade-off. Length robustness also differs: fine-tuned NMT degrades gradually as source sentences become longer, whereas LLMs show a sharper drop, losing up to 4 BLEU points from the third to the fifth length quantile. The high Spearman correlations among chrF, BERTScore, and COMET ($\rho \ge 0.81$) suggest that chrF may serve as a computationally cheaper proxy in preliminary experiments without sacrificing reliable model ranking.

The reliability of evaluating on a 500-sentence subset is supported by the close agreement between results on this subset and the full 5,000-sentence test set: NLLB-1.3B achieves BLEU 16.40 on the 500-sentence subset and 16.30 on the full set, a difference well within bootstrap confidence intervals. This indicates that our head-to-head comparisons involving LLMs, which were limited to 500 sentences, provide a fair and representative estimate of relative performance.

Although we did not conduct large-scale human evaluation, the neural metric COMET has been shown to correlate strongly with human judgments ($r > 0.9$) in WMT shared tasks~\citep{Rei2020}. Given that our findings are stable across multiple automatic metrics and supported by bootstrap significance testing, we consider the evidence robust.

In summary, this study benchmarks seven fine-tuned NMT models and four instruction-tuned LLMs on a large Arabic--Russian parallel corpus. The best fine-tuned model, NLLB-1.3B, achieves BLEU 16.3 and COMET 0.738, while all LLMs remain substantially below these results even in few-shot settings. Low lexical overlap is the dominant challenge, and morphological complexity disproportionately affects smaller models. Our results establish a solid baseline for the common low-resource scenario in which only a few thousand parallel sentences are available for training, and they demonstrate that, under such conditions, dedicated NMT models remain preferable to general-purpose LLMs for Arabic--Russian translation.

\section*{Limitations}

\textbf{Training data size.}
We fine-tuned all models on only 20k sentence pairs due to computational constraints. Although this size is realistic for low-resource settings, larger training sets would likely improve absolute translation quality and could alter the relative ranking of some models.

\textbf{Few-shot evaluation scale.}
Due to computational constraints, only the best-performing LLM was evaluated on the full 5,000-sentence test set; full-scale evaluation of all four LLMs remains future work, although the available result for Aya-Expanse confirms that the observed gap persists.

\textbf{Single reference translations.}
All automatic metrics rely on a single Russian reference for each source sentence. This setup can penalise valid lexical variation and may overestimate the proportion of \texttt{low\_overlap} errors, since legitimate paraphrases are not credited.

\textbf{Domain and dialect coverage.}
The Arabic--Russian Translation Corpus~\citep{ArabicNLPWorld:2026} is predominantly composed of Modern Standard Arabic from OPUS, with limited dialectal or domain-specific material. This distribution, while representative of currently available Arabic--Russian parallel data, limits the generalisability of our findings to dialectal Arabic or specialised domains.

\textbf{Error classification thresholds.}
The thresholds used for error categories (e.g., character-level Jaccard similarity for low overlap, length ratios for short and long outputs) were chosen heuristically based on the length distribution of the training data. These thresholds may require adjustment for other language pairs or domains.

\section{Future Work}

Several directions follow from this study. First, we plan to expand the corpus with domain-specific dictionaries covering medicine, engineering, law, and military terminology, and to scale fine-tuning to the full 15.8M-pair corpus to establish an upper-bound performance for this language pair. Second, explicit word-alignment and transliteration-based preprocessing may help reduce the lexical-overlap bottleneck identified in our error analysis. Third, we intend to evaluate newer multilingual models, including retrieval-augmented approaches, and to extend the benchmark to dialectal Arabic using resources such as MADAR~\citep{bouamor-etal-2018-madar}. Fourth, we plan to develop morphology-aware evaluation metrics tailored to Arabic--Russian, building on recent surveys of Arabic preprocessing and representation~\citep{alrekabee2025preprocessing}. Finally, we aim to validate our findings with targeted human evaluation on a sample of low-overlap and morphologically complex sentences.

\bibliographystyle{plainnat}
\bibliography{custom}

\end{document}